\documentclass[10pt,twocolumn,letterpaper]{article}

\usepackage{iccv}
\usepackage{times}
\usepackage{epsfig}
\usepackage{graphicx}
\usepackage{amsmath}
\usepackage{amssymb}
\usepackage{bm}
\usepackage{array}
\usepackage{tabu}
\usepackage{booktabs}
\usepackage{algorithmic}
\usepackage{multirow}

\usepackage[breaklinks=true,bookmarks=false]{hyperref}

\iccvfinalcopy 

\def\iccvPaperID{****} 

\ificcvfinal\fi
\begin{document}

\title{Multi-stage Multi-recursive-input Fully Convolutional Networks for Neuronal Boundary Detection}

\author{Wei Shen$^{1,2}$, Bin Wang$^1$, Yuan Jiang$^1$\thanks{\scriptsize equal contribution with the second author}, Yan Wang$^2$, Alan Yuille$^2$\\
$^1$ Key Laboratory of Specialty Fiber Optics and Optical Access Networks, Shanghai University\\
$^2$ Department of Computer Science, Johns Hopkins University\\
{\tt\footnotesize
wei.shen@t.shu.edu.cn,\{wangbin418,jy9387\}@outlook.com,\{wyanny.9,alan.l.yuille\}@gmail.com}}

\maketitle
\thispagestyle{empty}

\begin{abstract}
In the field of connectomics, neuroscientists seek to identify cortical connectivity comprehensively. Neuronal boundary detection from the Electron Microscopy (EM) images is often done to assist the automatic reconstruction of neuronal circuit. But the segmentation of EM images is a challenging problem, as it requires the detector to be able to detect both filament-like thin and blob-like thick membrane, while suppressing the ambiguous intracellular structure. In this paper, we propose multi-stage multi-recursive-input fully convolutional networks to address this problem. The multiple recursive inputs for one stage, i.e., the multiple side outputs with different receptive field sizes learned from the lower stage, provide multi-scale contextual boundary information for the consecutive learning. This design is biologically-plausible, as it likes a human visual system to compare different possible segmentation solutions to address the ambiguous boundary issue. Our multi-stage networks are trained end-to-end. It achieves promising results on two public available EM segmentation datasets, the mouse piriform cortex dataset and the ISBI 2012 EM dataset.
\end{abstract}

\section{Introduction} \label{sec:intro}
A central theme of neuroscience is to understand how a brain's functions are related to its neuronal structure~\cite{Ref:Lichtman11}. This is difficult because of the small size of neurons and the extremely high packing density of the neuropil, e.g., neurons densely packed axons and dendrites~\cite{Ref:Helmstaedter13}. Recent advances in high-throughput serial section electron microscopy (EM) have made possible the imaging of large volumes of neuronal tissue at high resolution, allowing neuroscience experts to reconstruct neuronal circuits and study the interconnections of neurons~\cite{Ref:Sporns05}. However, analysis of a large number of EM images by expert annotators is laborious and even impractical~\cite{Ref:Helmstaedter13}, which drives the demand for efficient automated neuronal circuit reconstruction approaches.

Serial section EM produces a stack of 2D images by cutting sections of brain tissue.
Due to the anisotropic resolutions of in-plane and out-of-plane, most neuronal circuit reconstruction approaches follow the following pipeline: (1) neuronal boundary detection on each 2D image, (2) neuronal structure segmentation based on the 2D boundary map, and (3) linking the neuronal segments across 2D images into a 3D reconstruction result.

\begin{figure}[!t]
\centering
\includegraphics[trim=4cm 12cm 1cm 3cm, clip=true, width=1.0\linewidth]{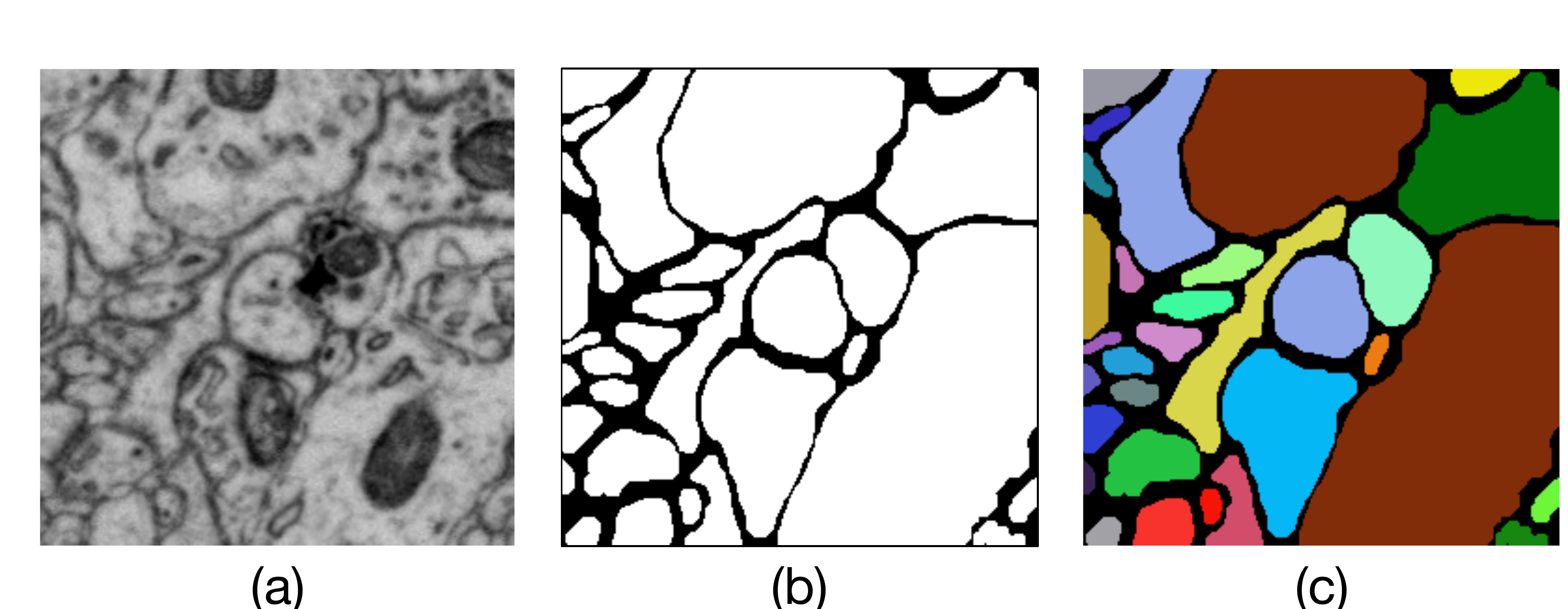}
\caption{Neuronal structure segmentation: an EM image (a) and the ground truths for its neuronal boundary detection result (b) and segmentation result (c), respectively.}\label{fig:img_edge_seg}
\end{figure}

This paper focuses on neuronal boundary detection on 2D images from serial section EM, also called membrane detection, which is the essential first step of the reconstruction pipeline. The challenges of this problem, as can be seen from Fig.~\ref{fig:img_edge_seg}, mainly lie in the following aspects: (1) The variation of the membranal thickness is large, as thin as a filament to as thick as a blob. (2) The noise of EM acquisition makes the membrane contrast to be low, inducing some membranal boundaries are even invisible. (3) The presence of confounding structures, such as mitochondria and vesicles, also increases the difficulty in membrane detection.

Driven by the rapid process of deep neural networks, more and more deep learning based methods have been proposed for neuronal boundary detection on EM images, achieving considerable progress~\cite{Ref:Ciresan12,Ref:Chen16,Ref:Fakhry17,Ref:Quan16,Ref:Lee15}. However, most of them only focus on how to improve detection performance by using the deep learning strategies shown to be effective on general computer vision problems, like image classification~\cite{Ref:Krizhevsky12,Ref:Simonyan14,Ref:He16}, semantic segmentation~\cite{Ref:Long15} and boundary/symmetry detection~\cite{Ref:Shen15,Ref:Xie15,Ref:Shen16}. For example, using fully convolutional networks enables holistic image training instead of patch-by-patch training~\cite{Ref:Chen16}; deeply supervised learning hierarchical representations~\cite{Ref:Chen16}; using residual structures rather than plain structures~\cite{Ref:Fakhry17,Ref:Quan16}. Although these strategies indeed improve membrane detection results, they lack an interpretation for this problem, i.e., how can membranes be detected and intracellular structures be suppressed meanwhile, even the contrasts of intracellular structures to context are much higher than those of membranes?

\begin{figure*}[!t]
\centering
\vspace{-1em}
\includegraphics[trim=0cm 4cm 0cm 1cm, clip=true, width=0.76\linewidth]{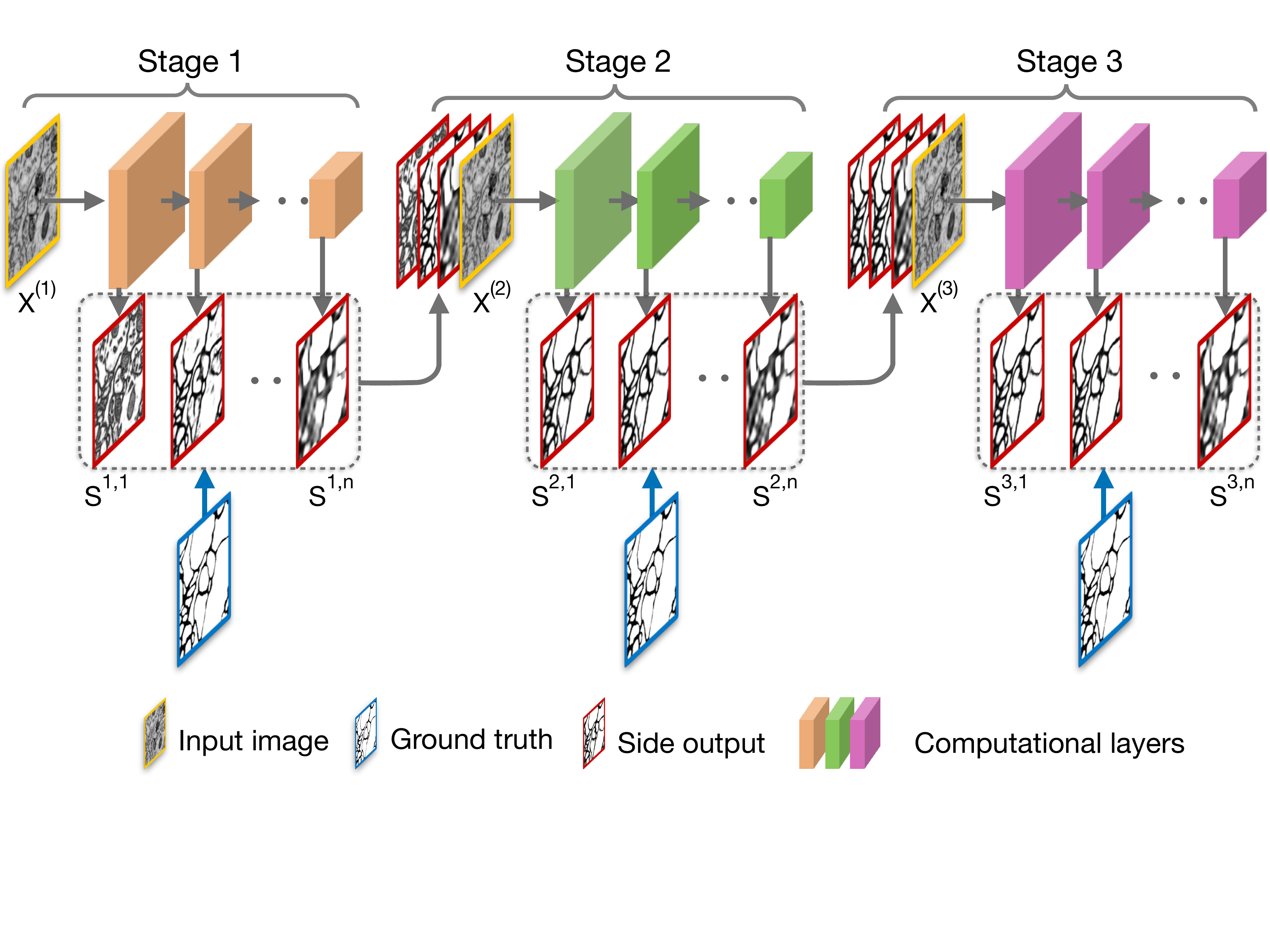}
\vspace{-1em}
\caption{The architecture of the proposed $\text{M}^2$FCN. The multiple side outputs of one stage are concatenated with the original image along the image channel
dimension, and are fed into the next stage. It makes the side outputs learned at the next stage approach ground truth.}\label{fig:network}
\end{figure*}

In this paper, we propose multi-stage multi-recursive-input fully convolutional networks ($\text{M}^2$FCN) for neuronal boundary detection. The architecture of $\text{M}^2$FCN is shown in Fig.~\ref{fig:network}. The whole net consists of multiple stages, where each stage generates multiple side outputs by imposing supervision at different levels~\cite{Ref:LeeXie15,Ref:Xie15} of the sub-net in this stage, and all these side outputs are concatenated with the original image to serve as the inputs for the next stage.

From a neurobiological prospective, this network architecture is biologically-plausible. First, as explained in~\cite{Ref:Lee15}, this recursive framework is in accord with the interplay process, named ``countercurrent disambiguating process''~\cite{Ref:Chen14}, between the primary and higher visual cortical areas (V1 and V4, respectively) in monkeys' brains, found by examining monkeys' performances on contour detection tasks. The latter stages of our networks act as V4 to detect the overall ``contour'' of neuronal boundaries and feed top-down influence to the early stages, which acts as V1, to enhance the activation on neuronal boundaries while suppressing those on intracellular structures. Second, as pointed out in~\cite{Ref:Helmstaedter13}, the ambiguous neuronal boundaries make segmentation difficult, and this issue can only be resolved when explicitly comparing the different possible segmentation solutions, which is what the human visual system may compute when inspecting this situation. We can roughly think that of the multiple recursive inputs, computed at the different levels in the previous stage, provide different possible segmentation solutions for training of the next stage. We show that using multiple recursive inputs is very important in our experiments, as it leads to much better results than using a single recursive input.

From the deep learning view, our networks take several advantages of the latest deep learning strategies to facilitate the learning of a neuronal boundary detection model, including (1) holistic image training and prediction benefited from using the architecture of fully convolutional neural networks, (2) supervised multi-scale feature learning at each level of each stage and (3) learning multi-stages in an end-to-end fashion, different from previous recursive approaches~\cite{Ref:Tu08,Ref:Lee15,Ref:Jurrus10,Ref:Dollar10} which learn a series of classifiers stepwise. Using multiple recursive inputs rather than one is a major difference between our networks and~\cite{Ref:Lee15}. Note that, the multiple recursive inputs for one stage are the multiple side outputs computed at the levels having different scales (receptive field sizes). In general, a side output with a small scale has better ability than one with larger scale for detecting a thin neuronal boundary between cluttered neurons. Conversely, a side output computed at a large scale can suppresses the false predictions on intracellular structures by using more image context. Therefore, these multiple recursive inputs provide richer information than one for the learning of next stage. We show that introducing these strategies in our networks can improve the performance of neuronal boundary detection.

We verify our networks on two public available EM segmentation datasets. One is the mouse piroform cortex EM dataset~\cite{Ref:Lee15}, a sizable and important EM dataset, contains 4 stacks of EM images of mouse piriform cortex, covering 460 images for training and 168 images for testing. We analyze alternative designs in our network architecture on this dataset and show it outperforms the state-of-the-arts~\cite{Ref:Lee15}. The other is the dataset used for ISBI 2012 EM segmentation challenge~\cite{Ref:Ronneberger15}, covering 30 images for training and 30 images for testing. Our networks can achieve a comparable result to the state-of-the art~\cite{Ref:Quan16,Ref:Drozdzal17} on this dataset.

In summary, our main contributions includes (1) end-to-end multi-stage networks architecture, in which each stage generates multiple side outputs by imposing supervision on different levels and feeds them into the next stage as the multiple recursive inputs. (2) using multiple recursive inputs rather than single recursive inputs can not only boost the performance of neuronal boundary detection, but also be biologically-plausible. (3) our networks achieve promising results on two public available EM segmentation datasets.

\section{Related Work}
Segmenting EM images of neural tissue is an important step to understand the circuit structure and the function of the brain \cite{Ref:Lee15,Ref:Quan16}. Early work of this topic needs to impose experts' knowledge. For example, users need to label intracellular regions to allow graph cut segmentation, and also correct segmentation errors afterwards \cite{Ref:Vu08}. To reduce the amount of human labor required~\cite{Ref:Helmstaedter13}, automatic neuron segmentation became an active research direction, which follows the pipeline that detects neuronal boundaries by machine learning algorithms~\cite{Ref:Laptev12,Ref:Kaynig10,Ref:Kumar10,Ref:Seyedhosseini11} and then applies post-processing algorithms, such as watershed~\cite{Ref:Najman96,Ref:Uzunbas14,Ref:Zlateski15}, hierarchical clustering~\cite{Ref:Nunez-Iglesias13,Ref:Liu14} and graph cut~\cite{Ref:Krasowski15} algorithms, to boundary maps to obtain neuron segments. But early methods, which are based on hand-crafted features,
tend to fail when the membrane is ambiguous.

With the rapid development of deep networks, segmentation and classification on medical images have undergone a vast revolution, from analyzing specific features of neuronal boundaries to designing fully automatic algorithm without experts' knowledge. Recent deep learning based neuron segmentation methods are well suited to deal with ambiguous neuronal boundaries in EM images. One of the earliest works made by Ciresan \emph{et al.} \cite{Ref:Ciresan12} used a succession of max-pooling convolutional networks as a pixel classifier, which estimated the probability of a pixel being a membrane. This method won the ISBI 2012 EM segmentation challenge \cite{Ref:Arganda-Carreras15}. Ronneberger \emph{et al.} \cite{Ref:Ronneberger15} presented a U-net structure with contracting paths, which captures multi-contextual information. Compared with the work in \cite{Ref:Ciresan12}, the U-net replaces pooling operations by upsampling operators, which propagates context information from multiple feature channels to higher resolution layers. Fully convolutional networks (FCNs) \cite{Ref:Long15} led to a breakthrough on semantic segmentation. With deep supervision and side outputs on FCN, a holistically-nested edge detector (HED) \cite{Ref:Xie15} was then proposed to solve the ambiguity in edge and object boundary detection. Due to the great success of these methods, Chen \emph{et al.} \cite{Ref:Chen16} presented a deeply supervised contextual network to fuse the multi-level side-outputs to segment the membrane. Deep residual network (ResNet) \cite{Ref:He16} addresses the degradation (of training accuracy) problem by optimizing the residual mapping, which ranked the 1st on ILSVRC 2015 image classification challenge~\cite{Ref:ILSVRC15}. This inspired fully residual convolutional neural networks with nested short and long skip connections for membrane segmentation, proposed by Quan \emph{et al.} \cite{Ref:Quan16} and Fakhry \emph{et al.} \cite{Ref:Fakhry17}. Most of the recent deep learning based methods were motivated by the novel network architectures proposed in the computer vision community. Although these methods achieve improvements on membrane detection, they lack interpretation and comprehensive analysis. Our proposed multi-stage multi-recursive-input fully convolutional networks ($\text{M}^2$FCN) not only considers the advantages of the latest deep network architectures but also is biologically plausible.

The recursive training framework has been applied to many computer vision tasks, such as image labeling~\cite{Ref:Tu08}, instance segmentation~\cite{Ref:Li16}, human pose estimation~\cite{Ref:Shen12}, and face alignment~\cite{Ref:Dollar10,Ref:Cao12}. However, they trained the recursive framework stepwise and only used one single recursive input. There are two image segmentation methods~\cite{Ref:Seyedhosseini13,Ref:SeyedhosseiniICCV13} also fed multi-recursive inputs into next stage in the recursive training framework. But, their strategies to generate the multi-recursive inputs are different from ours. In the first one~\cite{Ref:Seyedhosseini13}, the multi-recursive inputs for one stage were obtained by applying a series of Gaussian filters to the single output of the previous stage, but ours are the multiple outputs supervised at different levels in a deep network. The second one~\cite{Ref:SeyedhosseiniICCV13} downsampled an original input image into multiple input images with different resolutions and obtained the multi-recursive inputs from these multiple input images, but ours are computed from the same input image by using the hierarchy of a deep net. In addition, the multiple stages in their methods were trained in a stepwise manner. On the contrary, we embed the recursive learning in a deep network and first learn it in an end-to-end fashion.

Our work is related to the recursively trained network proposed in \cite{Ref:Lee15}, which trains a Very Deep 2D (VD2D) network first, then a Very Deep 2D-3D (VD2D3D) network initialized with learned 2D representations from VD2D network is trained to generate the boundary map. There are two important differences between the networks in \cite{Ref:Lee15} and our method. (1). We use multiple recursive inputs with different receptive field sizes to incorporate multi-level contextual boundary information learned from the previous stage, while VD2D3D only uses the single output of VD2D as its recursive input. (2). We train our network in an end-to-end fashion to co-enhance the learning ability (e.g., detect membranes while suppress intracellular structure) of all stages, while \cite{Ref:Lee15} sequentially learns deep networks. Benefited from end-to-end training and multiple recursive inputs, our networks can achieve better performance than VD2D3D while only using 2D EM images\vspace{-0.3em}.

\section{Multi-stage Multi-recursive-input Fully Convolutional Networks}
\subsection{Overview}
Fig.~\ref{fig:network} illustrates the proposed network architecture. Our networks consist of multiple sequential stages, where each stage is a sub-net built based on fully convolutional networks with multiple side outputs~\cite{Ref:Xie15}. Each sub-net consists of multiple levels, each of which is composed of several combinations of one convolutional layers followed by one ReLU layer and a final pooling layer. Each side output layer is connected to the last convolutional layer of each level, which is composed of a $1\times1$ convolutional layer and a deconvolutional layer to ensure the resolution of each side output is the same as the input original image. A sigmoid layer is applied to each side output layer to generate a neuronal boundary map having values belonging to $[0, 1]$.

Note that, the multiple side outputs of one stage are concatenated with the original image along the image channel dimension and fed into the next stage. By feeding multiple side outputs to the next stage, the side outputs of next stage can obtain the information from all scales, so that even low level side outputs in the next stage can capture large objects. While in a single stage network, each side output only corresponds to a certain scale, e.g., low level side outputs cannot suppress large intracellular structures, and high level side outputs cannot locate thin membranes accurately. This is like how a human visual system may compare different possible segmentation solutions and select the right scale from them \cite{Ref:Helmstaedter13}. The multiple stages in our networks can be trained in an end-to-end fashion, which enable the side outputs in a previous stage to receive feedback from those of the next stage, like V4 in the human  visual system gives the top-down influence to V1.
\subsubsection{Sub-net Architecture} \label{sec:sub-net}
We adopt the well-known HED network~\cite{Ref:Xie15} as our default sub-net, which is converted from VGG-16 net~\cite{Ref:Simonyan14}, having 5 levels, with strides 1, 2, 4, 8 and 16, respectively, and receptive field sizes 5, 14, 40, 92, 196, respectively. Each side output layer is connected to the last convolutional layer of each level, i.e., conv1\_2, conv2\_2, conv3\_3, conv4\_3, conv5\_3, respectively. 

\subsection{$\textbf{M}^2$FCN for Neuronal Boundary Detection}
Now we formulate our approach for neuronal boundary detection as a per-pixel classification problem. Given a raw input EM image $X=(x_j,j=1,\ldots,|X|)$, where index $j$ is over the image spatial dimensions of image $X$, the goal is to predict the neuronal boundary map $\hat{Y}=(\hat{y}_j,j=1,\ldots,|X|)$, where $\hat{y}_j\in\{0,1\}$ denotes the predicted label for each pixel $x_j$, i.e., if $x_j$ is predicted as a boundary pixel, $\hat{y}_j=0$; otherwise, $\hat{y}_j=1$. We learn an $\text{M}^2$FCN, which consists of $M$ stages and each stage has $N$ levels, to address this problem. Next, we introduce the training and testing phases of our approach respectively.
\subsubsection{Training Phase}
Since our networks use holistic image training, we consider each training image independently. Suppose that we are given a training batch with one 2D EM image $X$ and its corresponding ground truth neuronal boundary map $Y$, our goal is to supervise the multiple side outputs at different level to approach the ground truth map $Y$. Let $S^{m,n}=(s^{m,n}_j,j=1,\ldots,|X|)$ be the side output at the $n$-th level of the $m$-th stage. Since the side outputs of one stage will be fed to the next stage as the inputs, we express the input to the $m$-stage by
\small
\begin{equation}
X^{(m)}=X \oplus S^{m-1,1} \oplus \ldots \oplus S^{m-1,n}
\end{equation}
\normalsize
where $\oplus$ is the the concatenation operation along the image channel dimension and we define $X^{(1)}=X$.
Let $\mathbf{W}^m$ be the network parameters for the sub-net in the $m$-th stage and $\mathbf{w}^{m,n}$ be the parameters for the $n$-th side output in the $m$-th stage, we define a cross-entropy loss function for this side output by
\small
\begin{eqnarray}\label{eqn:loss_side_output}
&\ell^{m,n}(\mathbf{W}^m, \mathbf{w}^{m,n}; X^{(m)}, Y) =\nonumber\\
&-\beta\sum_{j \in |B|}\log (1-\sigma(s^{m, n}_j))\nonumber\\
&-(1-\beta)\sum_{j \in |\bar{B}|}\log (\sigma(s^{m,n}_j)).
\end{eqnarray}
\normalsize
This loss function (Eqn.\ref{eqn:loss_side_output}) is computed over all pixels in the training image $X$, where $|B|$ and $|\bar{B}|$ denote the boundary and non-boundary ground truth label sets respectively, $\sigma(\cdot)$ is the sigmoid function and $\beta=|\bar{B}|/|B|$ is a positive/negative class-balancing weight~\cite{Ref:Xie15} to eliminate the bias between boundary and non-boundary ground truths in training (in a typical 2D EM image, most of the ground truth is non-boundary). To supervise all the side outputs in our network, we define a loss function by
\small
\begin{eqnarray}
\mathcal {L}_s(\mathbf{W}, \mathbf{w}; X,Y)=\nonumber\\
\sum_{m=1}^M\sum_{n=1}^N\alpha_{m,n}\ell^{m,n}(\mathbf{W}^m, \mathbf{w}^{m,n}; X^{(m)}, Y),
\end{eqnarray}
\normalsize
where $\mathbf{W}=(\mathbf{W}^m,m=1,\ldots,M)$, $\mathbf{w}=(\mathbf{w}^{m,n},m=1,\ldots,M,n=1\ldots,N)$ and $\alpha_{m,n}$ is a loss weight for each side output.

To obtain a fused output $S^{f,m} $ for $m$-stage , we use a $1\times1$ convolutional layer to fuse its side outputs:
\small
\begin{equation}
S^{f,m} = \sum_{n=1}^Nh_{m,n}S^{m,n},
\end{equation}
\normalsize
where $\mathbf{h}=(h_{m,n},m=1,\ldots,M,n=1\ldots,N)$ is the fusion weight. Similar to Eqn.~\ref{eqn:loss_side_output}, a class-balanced cross-entropy loss function is defined for the fused output of $m$-stage:
\small
\begin{eqnarray}
\ell^{f,m}(\mathbf{W}^m, \mathbf{w}^{m}; X^{(m)}, Y)=\nonumber\\-\beta\sum_{j \in |B|}\log (1-\sigma(s^{f,m}_j)) -(1-\beta)\sum_{j \in |\bar{B}|}\log (\sigma(s^{f,m}_j)).
\end{eqnarray}
\normalsize
where $\mathbf{w}^m=(\mathbf{w}^{m,n},n=1\ldots,N)$. The loss function for all the fused outputs is
\small
\begin{eqnarray}
\mathcal {L}_f(\mathbf{W}, \mathbf{w}, \mathbf{h}; X,Y)=\nonumber\\
\sum_{m=1}^M\alpha_{f,m}\ell^{f,m}(\mathbf{W}^m, \mathbf{w}^{m}; X^{(m)}, Y),
\end{eqnarray}
\normalsize
where $\alpha_{f,m}$ is a loss weight for each stage.

All these parameters, $\mathbf{W}, \mathbf{w}, \mathbf{h}$, are optimized simultaneously by standard back-propagation:
\small
\begin{eqnarray}
(\mathbf{W}, \mathbf{w}, \mathbf{h})^{\ast} = \arg\min(\mathcal {L}_s(\mathbf{W}, \mathbf{w}; X,Y) \nonumber\\ +  \mathcal {L}_f(\mathbf{W}, \mathbf{w}, \mathbf{h}; X,Y)).
\end{eqnarray}
\normalsize
\subsubsection{Testing Phase}
In the testing stage, given an image $X$, by considering the $m$-th sub-net as a function $F_m$, we sequentially obtain the side outputs of the $m$-th sub-net by
\small
\begin{eqnarray}
(S^{m,n},n=1\ldots,N)=F_m(X^{(m)}, \nonumber\\ (\mathbf{W}^{1})^{\ast},\ldots,(\mathbf{W}^{m})^{\ast},(\mathbf{w}^{m,1})^{\ast},\ldots,(\mathbf{w}^{m,N})^{\ast}).
\end{eqnarray}
\normalsize
The fusion output of $m$-stage is obtained by
\small
\begin{equation}
S^{f,m}= \sum_{n=1}^Nh^{\ast}_{m,n}S^{m,n}.
\end{equation}
\normalsize
We use the fused output of the last stage $S^{f,M}$ as the final output of our network, then the unified boundary probability map is given by $\hat{Y}=\sigma(S^{f,M})$.
\subsubsection{Initialization for Multi-stage Training}
The proposed network is very deep (our 3-stage network with 5 levels has totally 48 layers). It is known that, training such a deep network from scratch is not easy.
Here, we adopt a simple strategy to initialize the multi-stage network. First, we train a single stage network. Then, this network is used to initialize the first stage of our network, while the rest of the network is randomly initialized. The initialization for the first stage provides high-confidence recursive inputs to the consecutive stage, which facilitates the training procedure.

\section{Experimental Results}
In this section, we discuss our designs for network architectures and training strategies, and compare our performance with other competitors.
\subsection{Experiment Setting}

The hyper parameters of our networks include: the mini-batch size (1), the loss weight for each side-output (1), the momentum (0.9), and the weight decay ($2\times10^{-4}$). We set the base learning rate to 1e-8 and pre-train a single stage network by 20,000 iterations. Then we use this pre-trained single stage network to initialize the first stage of our multi-stage network, and reduce the base learning rate to 1e-9 and train it by 10,000 iterations.
\subsection{Mouse Piriform Cortex Dataset}
The images of mouse piriform cortex dataset \cite{Ref:Lee15} were collected from the piriform cortex of an adult mouse, which contains 4 stacks of EM images.
We use the same training-testing split in \cite{Ref:Lee15}, i.e., stack2, stack3 and stack4 are for training and stack1 is for testing.


\subsubsection{Evaluation Metric}\label{sec:metric_piriform}
To evaluate membrane segmentation performances, we follow the protocol used in \cite{Ref:Lee15}, where the segmented membrane is measured by the Rand F-score:
\small
\begin{equation}
V_{Fscore}^{Rand} = \frac{2V_{merge}^{Rand}V_{split}^{Rand}}{V_{merge}^{Rand}+V_{split}^{Rand}},
\end{equation}
\normalsize
where $V_{merge}^{Rand}$ and $V_{split}^{Rand}$ are Rand merge score and Rand split score respectively, and defined by:
\small
\begin{equation}
V_{merge}^{Rand}=\frac{\sum_{ij}n_{ij}^2}{\sum_i(\sum_jn_{ij})^2}, \quad V_{split}^{Rand}=\frac{\sum_{ij}n_{ij}^2}{\sum_j(\sum_in_{ij})^2},
\end{equation}
\normalsize
where $n_{ij}$ denote the number of voxels in the $i$-th segment of the proposal segmentation and $j$-th segment of the ground truth segmentation. $V_{merge}^{Rand}$ and $V_{split}^{Rand}$ are close to $1$ when there are fewer merge and split errors, respectively. To calculate the Rand score, we obtain the neuronal segmentation based on the boundary map by applying the same modified graph-based watershed algorithm~\cite{Ref:Zlateski15} as in \cite{Ref:Lee15}. We use the default setting in~\cite{Ref:Lee15} and report our best Rand F-score $V_{Fscore}^{Rand}$.

\subsubsection{Data Augmentation}
Data augmentation is a standard way to generate sufficient training data for learning a ``good'' deep network. We rotate the images to 4 different angles ($0^\circ$, $90^\circ$, $180^\circ$, $270^\circ$) and flip them with different axis (up-down, left-right, no flip), then resize images to 3 different scales ($0.8$, $1.0$, $1.2$), totally leading to an augmentation factor of 36.

\subsubsection{Alternative Design Discussion}
\begin{figure*}[!ht]
\centering
\vspace{-1em}
\includegraphics[width=0.8\linewidth]{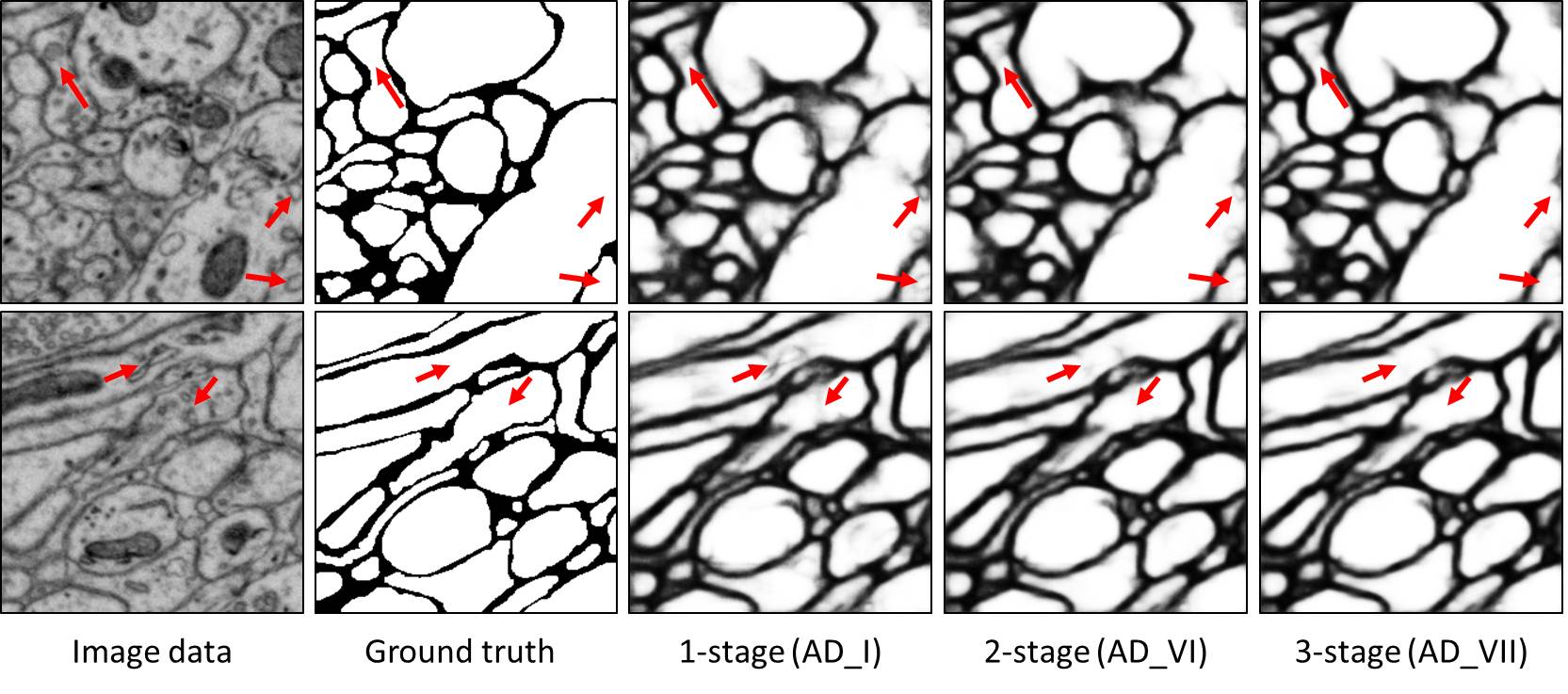}
\vspace{-0.5em}
\caption{Qualitative comparison between 1-stage (AD\_I), 2-stage (AD\_VI) and 3-stage (AD\_VII) networks. Red arrows indicate suppressed false detections by training more stages.}\label{fig:stage_comparison}
\end{figure*}

\begin{figure}[!ht]
\centering
\includegraphics[trim=14.5cm 8cm 0cm 0cm, clip=true, width=0.87\linewidth]{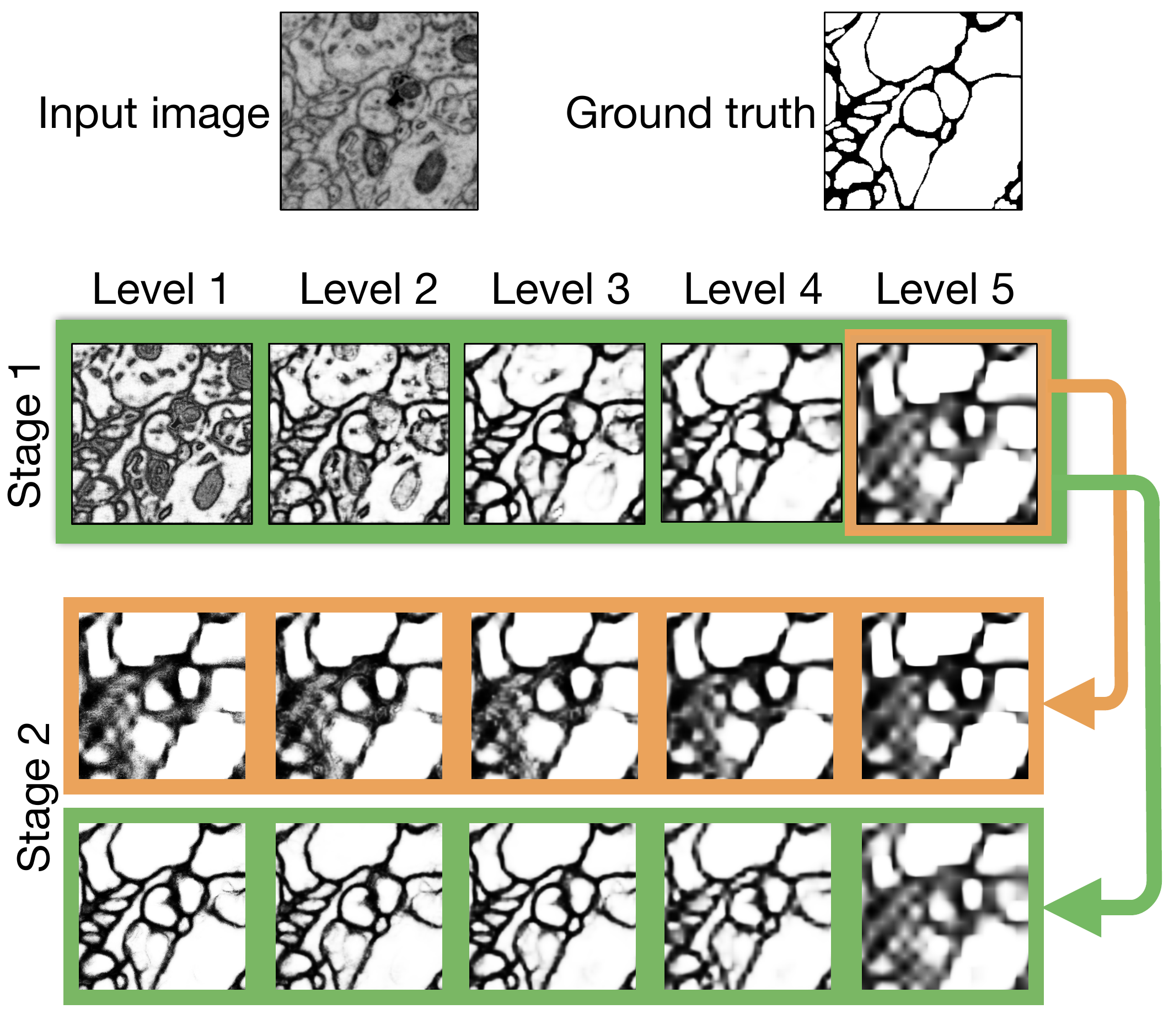}
\caption{The side outputs of the second stages learned by single recursive input (AD\_III) and multiple recursive inputs (AD\_VI).}\label{fig:multi-single}
\end{figure}

\begin{figure}[!ht]
\centering
\includegraphics[width=1.0\linewidth]{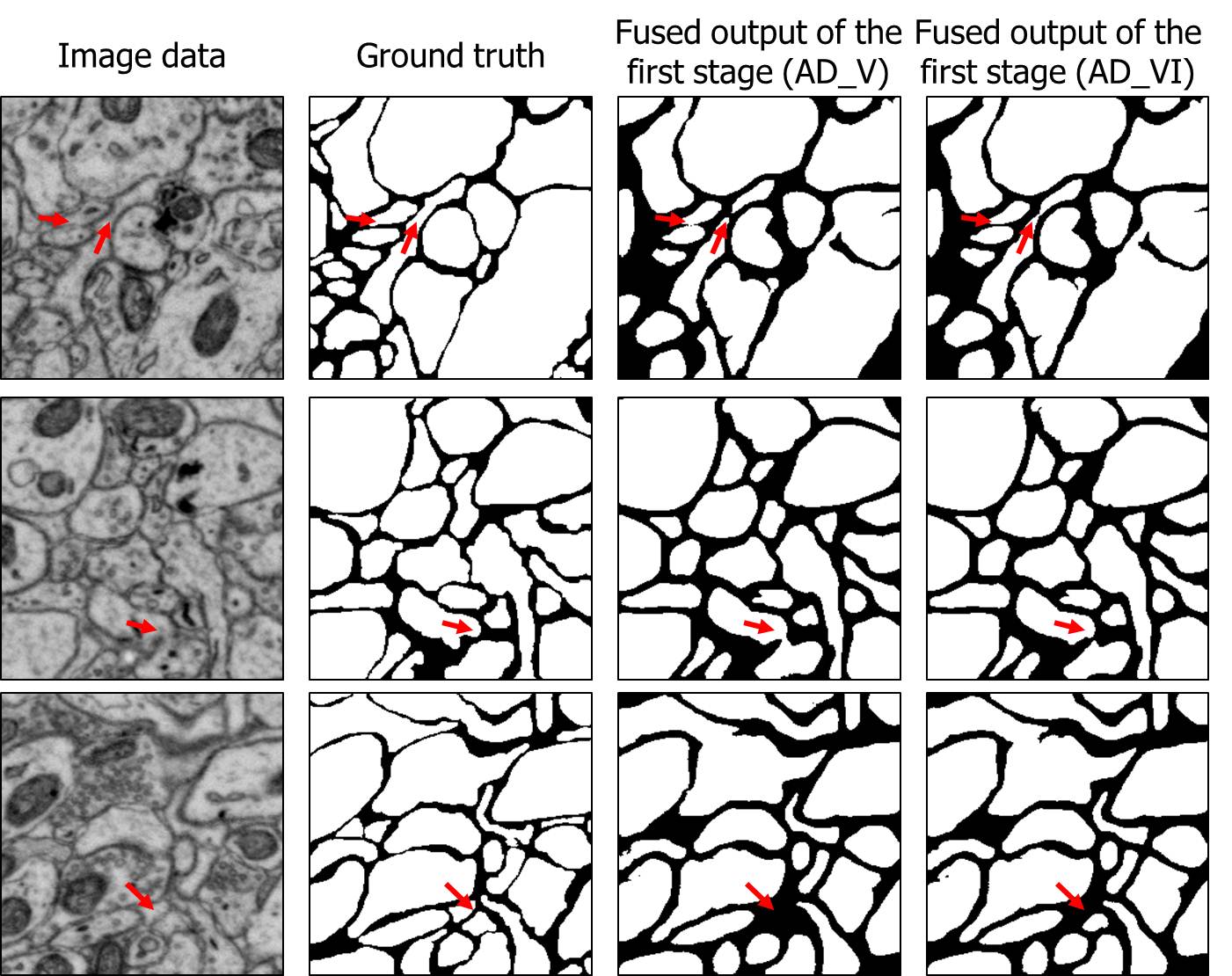}
\caption{The comparison between the fused outputs (after watershed~\cite{Ref:Zlateski15}) of the first stages of a 2-stage network trained stepwise (AD\_V) and end-to-end (AD\_VI). The latter one leads to better segmenting results (indicated by red arrows).}\label{fig:end_stepwise}
\end{figure}

We use the pre-trained single stage network as the baseline and discuss some possible alternative designs for network architectures and training strategies. The results of these alternative designs are summarized in Table.~\ref{tbl:alternative_design}. To simplify description, we denote each alternative design by ``AD'' plus an index.
\paragraph{The role of multiple stages.} Since our networks consist of multiple stages, it's necessary to see whether the performance can be improved by adding stage by stage. Due to the limitation of GPU memory, the deepest network we can train is a 3-stage network with sub-nets of 5 levels (Sec. \ref{sec:sub-net}). We compare the performance between a 1-stage network (AD\_I), a 2-stage network (AD\_VI) and a 3-stage network (AD\_VII). Note that, the AD\_I has the same architecture as the pre-trained single stage network. As shown in Table.~\ref{tbl:alternative_design}, the performance of AD\_I is almost the same as that of the pre-trained single stage network, which indicates that training a single stage network by more iterations cannot improve the performance considerably. The AD\_VI and the AD\_VII achieve $1.39\%$ and $1.86\%$ performance improvements compared with the baseline, respectively, which shows that our multi-stage training is effective for neuronal boundary detection. A qualitative comparison between these three networks, i.e, 1-stage (AD\_I), 2-stage (AD\_VI) and 3-stage (AD\_VII) networks, is given in Fig.~\ref{fig:stage_comparison}. Observed that, the false detections on intracellular structures such as mitochondria and vesicles can be reduced (indicated by red arrows) by training a network with more stages.
\paragraph{The role of multiple recursive inputs.}
As we stated in the Sec.~\ref{sec:intro}, using multiple recursive inputs for the multiple stage training is crucial for our framework. To evaluate this, we test two alternative network architectures, which only uses the side output of the 5-th level (AD\_III) and the 4-th level (AD\_IV) in each stage as the recursive input for the next stage respectively. As shown in Table.~\ref{tbl:alternative_design}, only using single recursive input results in a significant performance drop, $4.56\%$ decrease for AD\_III and $2.1\%$ decrease for AD\_IV. We illustrate the side outputs of the second stages of AD\_III and AD\_VI in Fig.~\ref{fig:multi-single}, where we see that the side outputs of the former one only response to large scale objects, while even the side outputs of the latter one can capture objects of different scales, thanks to the multiple recursive inputs.
\paragraph{Stepwise or end-to-end training.}
One difference between our multi-stage training framework to others~\cite{Ref:Tu08,Ref:Lee15,Ref:Jurrus10,Ref:Dollar10} is we train these multiple stages in an end-to-end fashion, not stepwise. To verify which way is better, we also train a 2-stage network stepwise, by simply fixing the parameters of the first stage in this 2-stage network (AD\_V). As can be seen from Table.~\ref{tbl:alternative_design}, training the 2-stage network stepwise leads to a considerable performance decrease ($0.9819\rightarrow0.9762$). During end-to-end training, previous stages are influenced by next stages. We visualize the fused outputs (after watershed) of the first stages of (AD\_V) and (AD\_VI), respectively, in Fig.~\ref{fig:end_stepwise}, where we see the latter one leads to better segmentation results (indicated by red arrows). Therefore, we conclude that training in an end-to-end way is better.
\paragraph{The range of the receptive field sizes in each sub-net.}
The receptive field sizes of the 5 levels in each sub-net range from 5 to 196. Such a wide range of receptive field sizes ensure the side outputs to be able to capture small neuronal boundaries while suppress relative big intracellular structures. To show this wide range of receptive field sizes is important for neuronal boundary detection, we use an alternative network architecture for each sub-net, which is obtained by removing the last level from the default sub-net, i.e., a 2-stage network with 4-level sub-net (AD\_II). As we expected, as shown in Table.~\ref{tbl:alternative_design}, removing the last level in each sub-net leads to performance decrease.
\begin{table}[!th]
\centering
\caption{Performance of alternative designs for network architectures and training strategies.}\label{tbl:alternative_design}
\begin{tabu}{l|c}
Alternative Design & $V_{Fscore}^{Rand}$\\
\tabucline[2pt]{-}

pre-trained single stage, 5-level (baseline) &{0.9680}\\
\hline
AD\_I: 1-stage, 5-level &{0.9688}\\
\hline
AD\_II: 2-stage, 4-level, end-to-end, &\multirow{2}{*}{0.9739}\\
multi-recursive-input & $$\\
\hline
AD III: 2-stage, 5-level, end-to-end,&\multirow{2}{*}{0.9410}\\
single-recursive-input (level 5) & $$\\
\hline
AD\_IV: 2-stage, 5-level, end-to-end,&\multirow{2}{*}{0.9656}\\
single-recursive-input (level 4) & $$\\
\hline
AD\_V: 2-stage, 5-level, stepwise, &\multirow{2}{*}{0.9762}\\
multi-recursive-input & $$\\
\hline
AD\_VI: 2-stage, 5-level, end-to-end,&\multirow{2}{*}{0.9819}\\
multi-recursive-input & $$\\
\hline
AD\_VII: 3-stage, 5-level, end-to-end,& \multirow{2}{*}{\textbf{0.9866}}\\
multi-recursive-input & $$\\
\hline
\end{tabu}
\end{table}
\subsubsection{Performance Comparison}
\begin{figure}[!ht]
\centering
\includegraphics[trim=3cm 7cm 4cm 7.4cm, clip=true, width=0.8\linewidth]{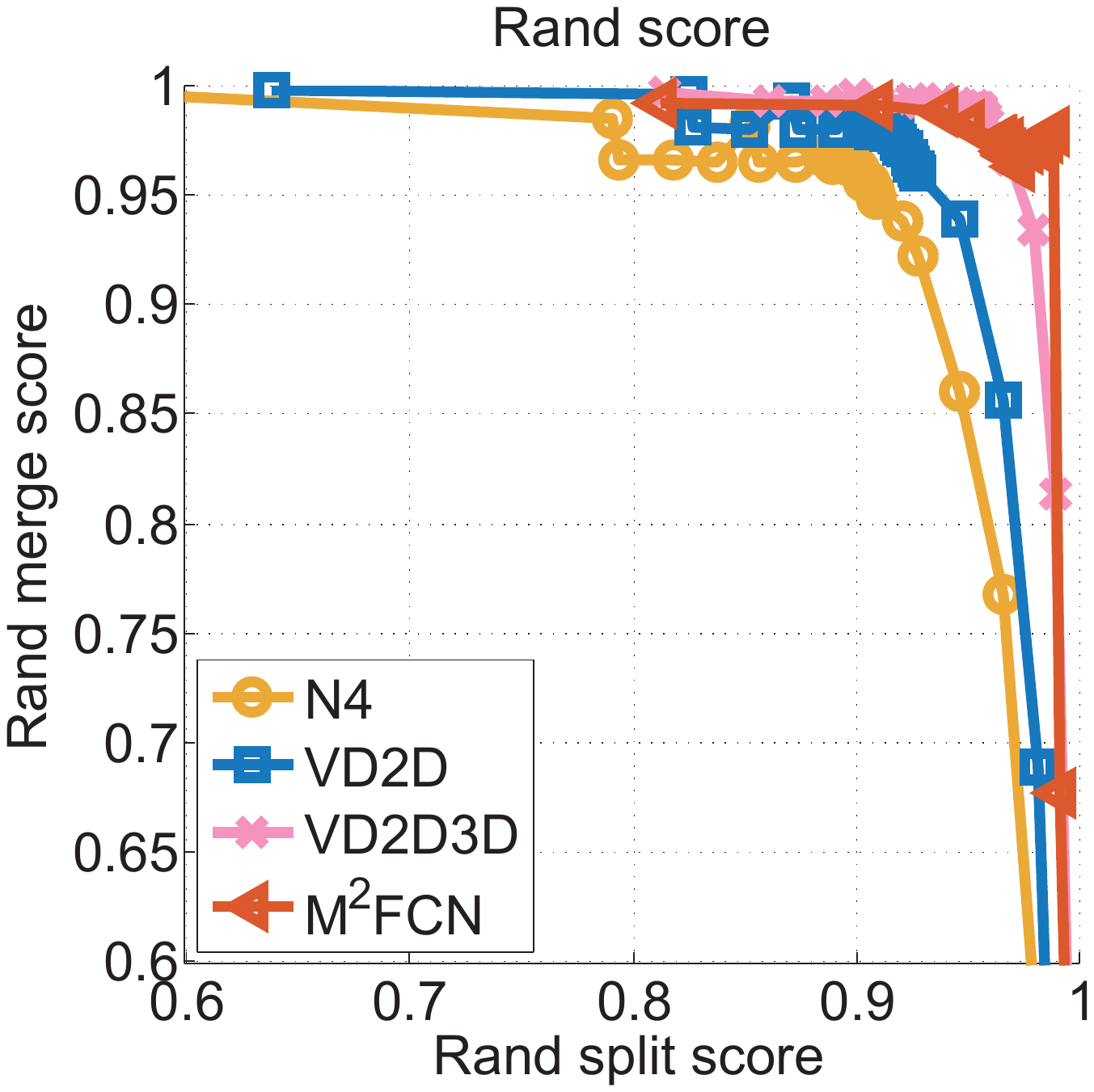}
\vspace{-1em}
\caption{Evaluation of neuronal boundary detection methods by precision (rand merge)-recall (rand split) curves.}\label{fig:rand_score_curve}
\end{figure}

\begin{figure}[!ht]
\centering
\includegraphics[width=1.0\linewidth]{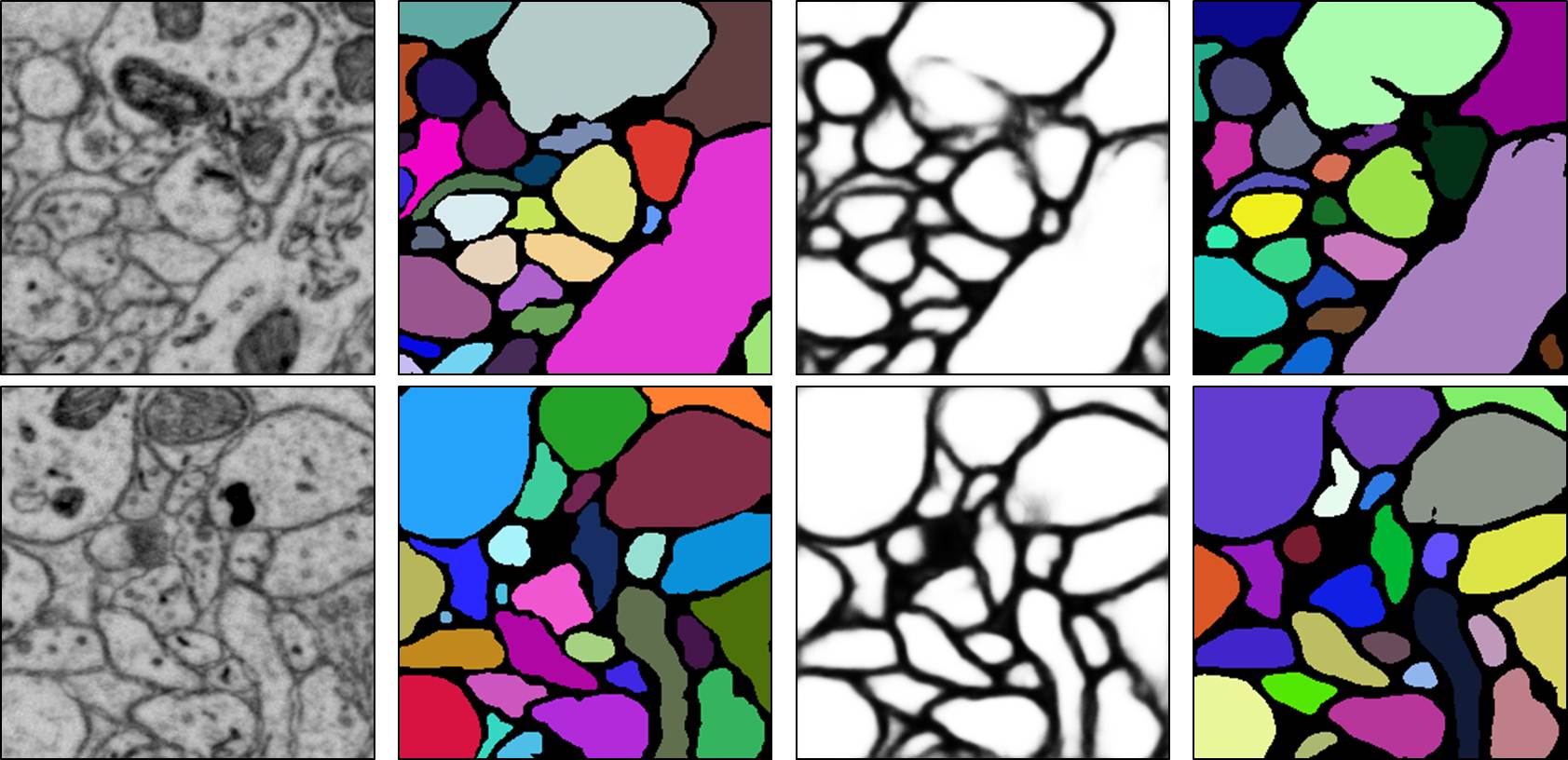}
\caption{Neuron segmentation examples. From left to right: the original image, the ground truth segmentation, our boundary map and the segmentations obtain by applying the watershed algorithm~\cite{Ref:Zlateski15} to our boundary maps.}\label{fig:segmentation}
\end{figure}

Now we compare our networks (3-stage with 5 levels) with other competitors on the mouse piriform cortex dataset. The quantitative results are summarized in Table~\ref{tbl:rand_f_score} and the precision (rand merge)-recall (rand split) curves are illustrated in Fig.~\ref{fig:rand_score_curve}. As can be seen, our method can maintain a high precision even when it achieves a high recall, thanks to the multi-stage training which suppresses false detections of boundaries on intracellular structures while enhancing neuronal boundaries. The current state-of-the-art method on the mouse piriform cortex dataset is VD2D3D~\cite{Ref:Lee15}, which is also a recursive training framework. But it trains two stages stepwise, i.e., train the first one, a 2D convolutional network, then uses its output as the recursive input for the second one, a 3D convolutional networks. The experimental results show that with the end-to-end multi-stage training and multi-recursive-inputs, our 2D 2-stage network can achieve better performance than a 2D-3D network. Note that, as VD2D3D already obtained a high Rand F-score, our method achieves around $1.5\%$ improvement on it, which is meaningful. Such a low error obtained on a large EM image dataset is important for neuron reconstruction. Some neuron segmentation results obtained by applying the watershed algorithm~\cite{Ref:Zlateski15} to our boundary maps are shown in Fig.~\ref{fig:segmentation}.

\begin{table}[!th]
\centering
\caption{Neuronal boundary detection performance comparison between different methods. The values of $V_{merge}^{Rand}$ and $V_{split}^{Rand}$ correspond to the best Rand F-score $V_{Fscore}^{Rand}$.}\label{tbl:rand_f_score}
\begin{tabular}{ccccccc}
\toprule
Method&$V_{merge}^{Rand}$&$V_{split}^{Rand}$&$V_{Fscore}^{Rand}$\\
\midrule
N4~\cite{Ref:Ciresan12}&0.9619&0.9010&0.9304\\
VD2D~\cite{Ref:Lee15}&0.9771&0.9174&0.9463\\
VD2D3D~\cite{Ref:Lee15}&0.9891&0.9555&0.9720\\
$\text{M}^2$FCN (1 stage)&0.9576&0.9802&0.9688\\
$\text{M}^2$FCN (2 stage)&0.9759&0.9880&0.9819\\
$\text{M}^2$FCN (3 stage)&0.9917&0.9815&\textbf{0.9866}\\
\bottomrule
\end{tabular}
\end{table}

\subsection{ISBI 2012 EM segmentation dataset}
Most of current neuronal boundary detection methods are evaluated on the public dataset of ISBI 2012 EM segmentation challenge~\cite{Ref:Ronneberger15}. The training data of this dataset is a set of 30 consecutive images (512 $\times$ 512 pixels) from a serial section Transmission Electron Microscopy (ssTEM) dataset of the Drosophila first instar larva ventral nerve cord~\cite{Ref:Cardona2010}. The testing data of this dataset also contains 30 consecutive EM images of the same resolution. The ground truth boundary maps of the training images are made publicly available to enable participants to develop their algorithm, while the ground truth boundary maps of the test images are kept by the organizers. Although this challenge is over, it is still open for submissions. The performance of the new submissions will be reported on the leader board of this challenge. There are over 70 results listed on the leader board, but not all of them have published papers. We summarized some leading quantitative results reported in published papers in Table~\ref{tbl:ISBI2012}. Note that, many state-of-the-art methods apply post-processing or average multiple trained models to boost the performance, such as PolyMtl~\cite{Ref:Drozdzal17}, FusionNet~\cite{Ref:Quan16} and CUMedVision~\cite{Ref:Chen16}. Our method, a two-stage network using a single trained model without post-processing, can achieve 0.9780 Rand F-score, which is comparable with the state-of-the-art methods and better than CUMedVision~\cite{Ref:Chen16}, a one-stage HED. But, CUMedVision used post-processing and averaged 6 trained models to improve the result. This comparison shows the effectiveness of our multi-stage learning. IAL IC~\cite{Ref:Beier16} is a post-processing method, which can be applied to our result to improve our performance. Besides, as both FusionNet and PolyMtl are built on ResNet~\cite{Ref:He16}, we can also replace the sub-net in our model by such a powerful network to gain improvement.

\begin{table}[!th]
\centering
\caption{Comparison to published entries on the ISBI 2012 EM dataset~\cite{Ref:Ronneberger15}. For full ranking of all submitted methods, please refer to the challenge website: \url{http://brainiac2.mit.edu/isbi_challenge/leaders-board-new}.}\label{tbl:ISBI2012}
\begin{tabular}{cc}
\toprule
Method&$V_{Fscore}^{Rand}$\\
\midrule
PolyMtl~\cite{Ref:Drozdzal17}&0.9806\\
$\text{M}^2$FCN (ours)&0.9780\\
FusionNet~\cite{Ref:Quan16}&0.9780\\
IAL IC~\cite{Ref:Beier16}&0.9773\\
CUMedVision~\cite{Ref:Chen16}&0.9768\\
FCN+LSTM~\cite{Ref:ChenNIPS16}&0.9754\\
Unet~\cite{Ref:Ronneberger15}&0.9727\\
\bottomrule
\end{tabular}
\end{table}
\section{Conclusion}
We present multi-stage multi-recursive-input fully convolutional networks for neuronal boundary detection. In the proposed architecture, the multiple side outputs learned at different scales in one stage, are fed into the next stage. This provides the ability to detect neuronal boundaries while suppressing false predictions on intracellular structures. Extensive analysis on two public EM segmentation datasets, the mouse piriform cortex dataset and the ISBI 2012 EM dataset, verifies the advantages of our network architecture.

\noindent {\bf Acknowledgement}. This work was supported in part by
the National Natural Science Foundation of China No.~61672336 and No.~61303095, in part by ``Chen Guang'' project supported by Shanghai Municipal Education Commission and Shanghai                                                        Education Development Foundation No.~15CG43, and in part by NSF CCF-1231216.
{\small
\bibliographystyle{ieee}
\bibliography{M2FCN}
}

\end{document}